\documentclass[final]{cvpr}

\usepackage{times}
\usepackage{epsfig}
\usepackage{graphicx}
\usepackage{amsmath}
\usepackage{amssymb}

\usepackage{multirow}
\usepackage{array}
\usepackage{caption}
\usepackage{makecell}

\usepackage[pagebackref=true,breaklinks=true,colorlinks,bookmarks=false]{hyperref}

\def\cvprPaperID{5128} 
\def\confYear{CVPR 2021}

\begin{document}

\title{Encoder Fusion Network with Co-Attention Embedding\\for Referring Image Segmentation}

\author{Guang Feng$^{1,2}$, Zhiwei Hu$^{1,2}$, Lihe Zhang$^{1,2,}$\footnotemark[2], Huchuan Lu$^{1,2}$\\
\small $^1$School of Information and Communication Engineering, Dalian University of Technology \\ \small $^2$Ningbo Institute, Dalian University of Technology\\
{\tt\small fengguang.gg@gmail.com, hzw950822@mail.dlut.edu.cn, }\\
{\tt\small \{zhanglihe, lhchuan\}@dlut.edu.cn}\\
}

\maketitle
\thispagestyle{empty}
\renewcommand{\thefootnote}{\fnsymbol{footnote}}
\footnotetext[2]{Corresponding Author}

\begin{abstract}
   Recently, referring image segmentation has aroused widespread interest. Previous methods perform the multi-modal fusion between language and vision at the decoding side of the network. And, linguistic feature interacts with visual feature of each scale separately, which ignores the continuous guidance of language to multi-scale visual features. In this work, we propose an encoder fusion network (EFN), which transforms the visual encoder into a multi-modal feature learning network, and uses language to refine the multi-modal features progressively. Moreover, a co-attention mechanism is embedded in the EFN to realize the parallel update of multi-modal features, which can promote the consistent of the cross-modal information representation in the semantic space. Finally, we propose a boundary enhancement module (BEM) to make the network pay more attention to the fine structure. The experiment results on four benchmark datasets demonstrate that the proposed approach achieves the state-of-the-art performance under different evaluation metrics without any post-processing.
\end{abstract}

\section{Introduction}

Referring image segmentation aims to extract the most relevant visual region (object or stuff) in an image based on the referring expression.
Unlike the traditional semantic and instance segmentation, which require to correctly segment each semantic category or each object in an image, referring image segmentation needs to find a certain part of the image according to the understanding of the given language query. Therefore, it can be regarded as a pixel-wise foreground/background segmentation problem, and the output result is not limited by the predefined semantic categories or object classes. This task has a wide range of potential applications in language-based human-robot interaction.

\begin{figure}[t]
\vspace{0mm}
\begin{tabular}{c@{}c}
\includegraphics[width=0.96\linewidth]{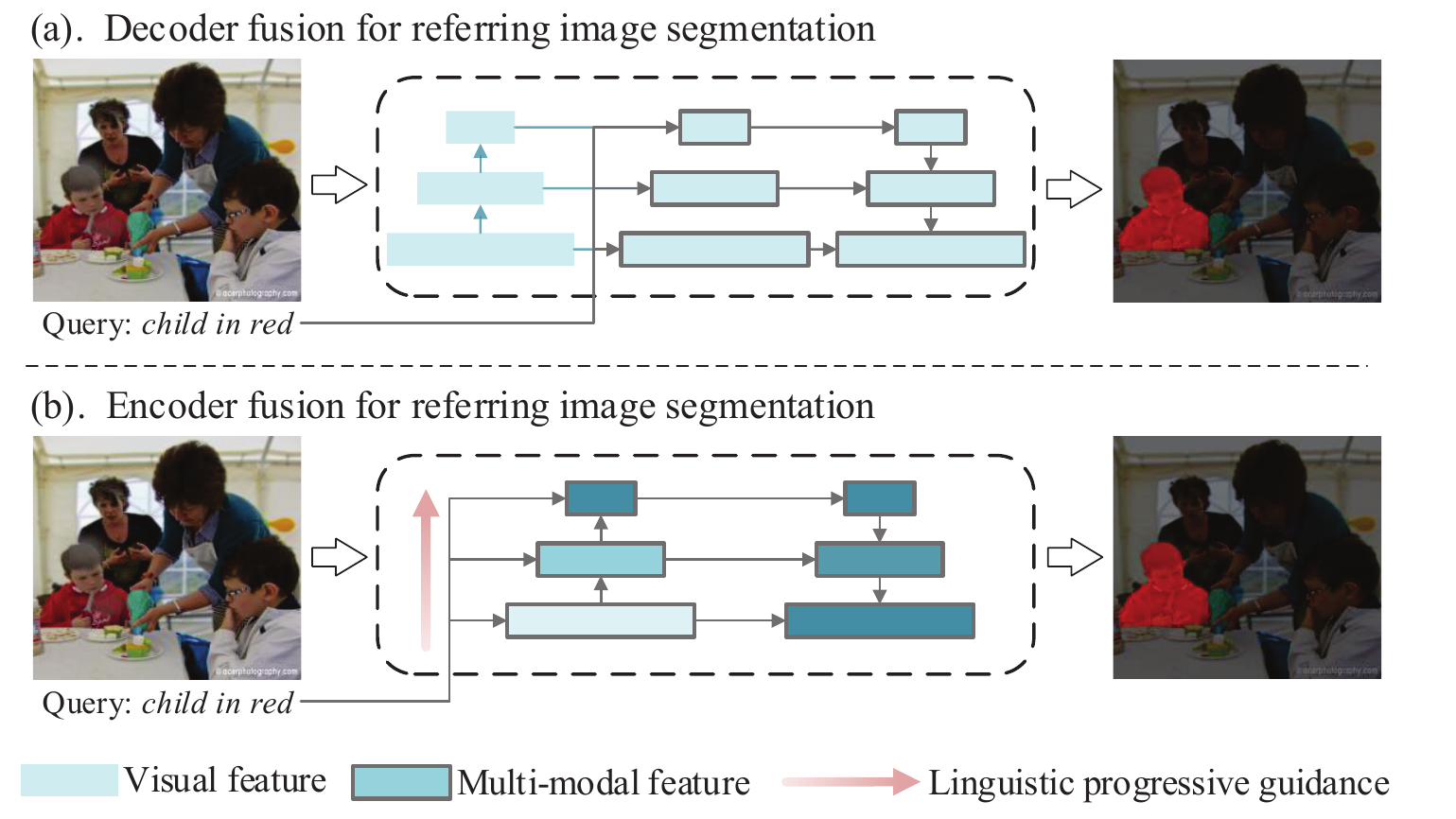}
\ \\
\end{tabular}
{\begin{center}
\vspace{-7mm}
\caption{\small{Two multi-modal fusion mechanisms.
Existing methods achieve the fusion between language and vision in the decoder, while the proposed method does it in the encoder.
}}
\label{fig:introduction}
\end{center}
}
\vspace{-9mm}
\end{figure}

The key of this task is to realize the cross-modal matching between visual and linguistic features. The deep-learning community has rapidly improved the results of vision-language tasks over a short period of time. The rapid development of convolutional neural network (CNN) and recurrent neural network (RNN) have made a qualitative leap in the ability of understanding vision and language, thereby they can solve more complex pixel-level cross-modal prediction tasks. Early referring image segmentation methods~\cite{hu2016segmentation,liu2017recurrent,li2018referring,margffoy2018dynamic} mainly rely on the powerful learning ability of deep learning model. They directly concatenate linguistic features with visual features of each region, and then use the combined multi-modal features to generate the segmentation mask. Due to the lack of sufficient interaction between two modalities, such solutions can not meet the requirements of real-world applications.
Recently, some works~\cite{shi2018key,ye2019cross,chen2019see,hu2020bi,huang2020referring,hui2020linguistic} began to consider the linguistic and visual attention mechanisms to better aggregate these two kinds of features.

Although some referring image segmentation methods have been proposed in the last few years, there are still many problems that have not been explored. On the one hand, for the cross-modal fusion of vision and language. Previous methods usually adopt the \textbf{decoder fusion} strategy, in which the RGB image and the referring expression are fed into CNN or RNN to generate their own feature representations separately, and then fuse these features in the decoding stage.
However, this fusion strategy at the output side of the network either only considers the interaction between linguistic and highest-level visual features~\cite{liu2017recurrent,li2018referring} or combines the linguistic features with the visual features of each level independently (as shown in Fig.~\ref{fig:introduction} (a))~\cite{ye2019cross,hu2020bi,hui2020linguistic}. They do not investigate the deep guidance of language to multi-modal fused features.
Besides, some works utilize visual and linguistic attention mechanisms for cross-modal feature matching. But they update the linguistic and visual features in a serial mode~\cite{shi2018key,chen2019see,hu2020bi,huang2020referring,hui2020linguistic}, that is, they only update the feature of one modality at a specific time, which will lead to the update delay of the features between different modalities and eventually weaken the consistency of the representation of multi-modal information. On the other hand, in CNNs, the repeated stride and pooling operations may lead to the loss of some important fine-structure information, but few referring image segmentation methods explicitly consider the problem of detail recovery.

To resolve the aforementioned problems, we propose an encoder fusion network with co-attention embedding (CEFNet) for referring image segmentation. Instead of the cross-modal information fusion at the output side, we adopt the \textbf{encoder fusion} strategy for the first time to progressively guide the multi-level cross-modal features by language. The original visual feature encoder (e.g., ResNet) is transformed into a multi-modal feature encoder (as shown in Fig.~\ref{fig:introduction} (b)). The features of two modalities are deeply interleaved in the CNN encoder.
Furthermore, to effectively play the guiding role of language, we adopt the co-attention mechanism to simultaneously update the features of different modalities. It utilizes the same affinity matrix to project different features to the common feature subspace in a parallel mode and better achieve the cross-modal matching to bridge the gap between coarse-grained referring expression and highly localized visual segmentation.
We implement two simple and effective co-attention mechanisms such as vanilla co-attention and asymmetric co-attention, which offer a more insightful glimpse into the task of referring image segmentation. Finally, we design a boundary enhancement module (BEM), which captures and exploits boundary cues as guidance to gradually recover the details of the targeted region in the decoding stage of the network.

Our main contributions are as follows:
\vspace{0mm}
\begin{itemize}
\vspace{-0mm}
\item We propose an encoder fusion network (EFN) that uses language to guide the multi-modal feature learning, thereby realizing deep interweaving between multi-modal features. In the EFN, the co-attention mechanism is embedded to guarantee the semantic alignment of different modalities, which promotes the representation ability of the language-targeted visual features.
\vspace{-2mm}
\item We introduce a boundary enhancement module (BEM) to emphasize the attention of the network to the contour representation, which can help the network to gradually recover the finer details.
\vspace{-0mm}
\item The proposed method achieves the state-of-the-art performance on four large-scale datasets including the UNC, UNC+, Google-Ref and ReferIt with the speed of 50 FPS on an Nvidia GTX 1080Ti GPU.
\end{itemize}

\section{Related Work}
\begin{figure*}[t]
\centering
\begin{tabular}{c@{}c}
\includegraphics[width=0.96\linewidth]{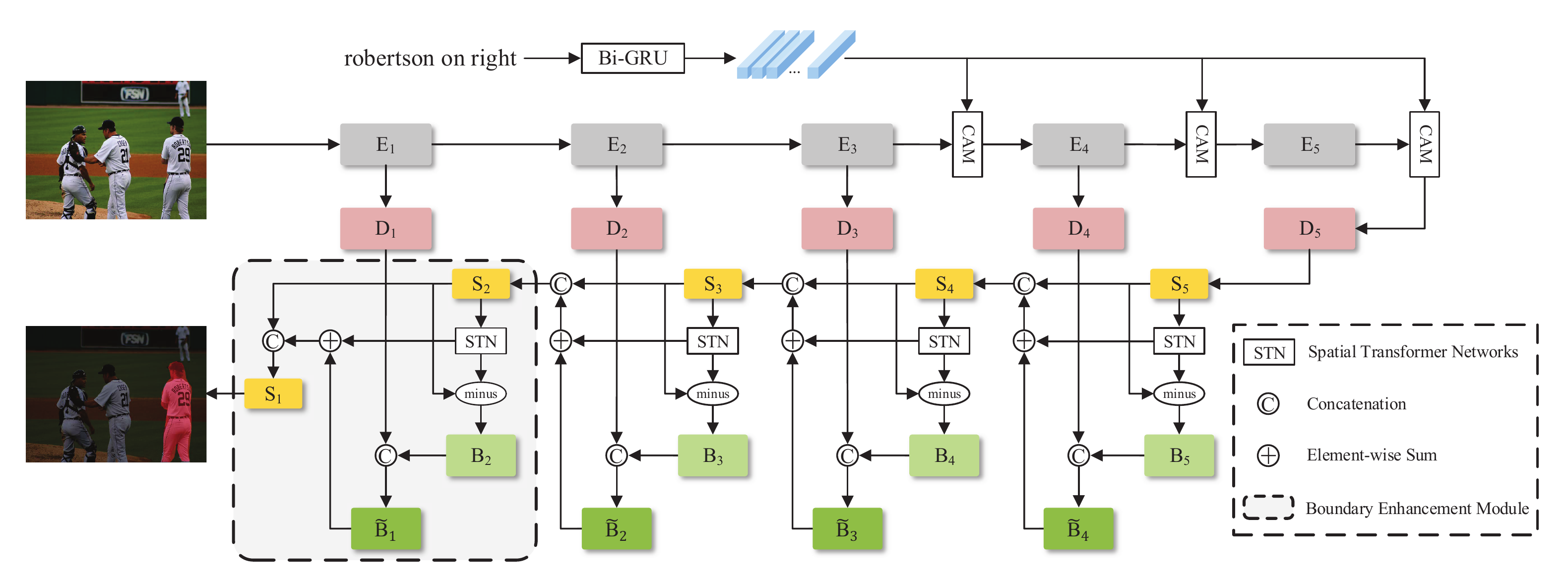}
\ \\
\end{tabular}
{\begin{center}
\vspace{-8mm}
\caption{\small{The overall architecture of our model. It mainly consists of the Bi-GRU encoder, ResNet-101 encoder ($E_1 \sim E_5$), co-attention module (CAM), decoder blocks ($D_1 \sim D_5$) and boundary enhancement module (BEM) assembled at the decoding end. The CAM is used to realize the matching between multi-modal features. The BEM captures the boundary cues and uses them to recover the details of the image, which produces a more accurate segmentation mask. The details of the proposed method are introduced in Sec.~\ref{sec:p_method}}}
\label{fig:structure}
\end{center}
}
\vspace{-6mm}
\end{figure*}
\noindent\textbf{Semantic and Instance Segmentation.}
The former aims at grouping pixels in a semantically meaningful way without differentiating each instance.
The latter requires to separate all instances of objects rather than the stuff.
In recent years, many semantic segmentation methods adopt fully convolutional network (FCN)~\cite{long2015fully} for end-to-end prediction.
On this basis, the multi-scale context~\cite{zhao2017pyramid,chen2017deeplab,chen2017rethinking,ding2018context,feng2020cacnet} and attention mechanisms~\cite{zhao2018psanet,fu2019dual,zhu2019asymmetric,huang2019ccnet}
are deeply examined.
Some works~\cite{noh2015learning,ding2018context} leverage the encoder-decoder structures to alleviate the loss of details caused by continuous down-sampling.
Also, RGB-D based methods~\cite{hazirbas2016fusenet,chen2020bi} introduce depth prior to improve the performance. These methods
provide inspirations for referring image segmentation.

In instance segmentation, Mask-RCNN~\cite{he2017mask} is a classical framework, which uses two-stage design to sequentially generates proposals and classify/segment them.
In the follow-up works, feature pyramid~\cite{lin2017feature}, top-down and bottom-up~\cite{liu2018path}, iterative optimization~\cite{chen2019hybrid} and boundary-aware mechanisms~\cite{cheng2020boundary} are explored.
The success of boundary refinement strategy provides us with an important clue to solve the problem of referring image segmentation.

\noindent\textbf{Referring Image Comprehension.}
This task has two branches: localization and segmentation.
For referring image localization, previous methods are mainly composed of two separate stages. They firstly use object detector to extract candidate regions, and then rank these regions according to the referring expression.
Pioneering methods~\cite{hu2016natural,mao2016generation,luo2017comprehension} use the CNN-LSTM structure to select the object with the maximum posterior probability of the expression, and other works~\cite{liu2017referring,yu2017joint} optimize the joint probability of the target object and the expression.
Recently, some methods~\cite{yang2019fast,sadhu2019zero,liao2020real} use a one-stage framework. Instead of generating  excessive candidate boxes, they directly predict the coordinates of the targeted region in an end-to-end manner. The above methods all implement the multi-modal fusion in the decoder.

For referring image segmentation, early methods~\cite{hu2016segmentation,liu2017recurrent,li2018referring,margffoy2018dynamic} directly concatenate language and visual features and then completely depend on a fully convolutional network to infer the pixel-wise mask. These methods do not explicitly formulate the intra-modal and inter-modal relationships. Some recent works~\cite{shi2018key,ye2019cross,chen2019see,hu2020bi,huang2020referring,hui2020linguistic} consider the self-attention and cross-attention mechanisms of linguistic and visual information.
For example, Shi et al.~\cite{shi2018key} adapt the vision-guided linguistic attention to learn the adaptive linguistic context of each visual region.
Ye et al.~\cite{ye2019cross} employ multiple non-local modules to update each pixel-word mixed features in a fully-connected manner.
Hu et al.~\cite{hu2020bi} design a bi-directional relationship inferring network to model the relationship between language and vision, which realizes the serial mutual guidance between multi-modal features.
Huang et al.~\cite{huang2020referring} firstly perceives all the entities in the image according to the entity and attribute words, and then use the relational words to model the relationships of all entities.
LSCM~\cite{hui2020linguistic} utilizes word graph based on dependency parsing tree to guide the learning of multi-modal context.
Similarly, these methods also use the decoder fusion strategy.
In addition, they do not update linguistic and visual features in parallel, which may weaken the consistency of language and vision in the semantic space. Different from the previous works, we design a parallel update mechanism to enhance the compatibility of multi-modal representation, and the multi-modal feature matching is performed in the encoder. We also propose a boundary enhancement module to guide the progressive fusion of multi-level features in the decoding stage.
\section{Proposed Method}
\label{sec:p_method}
The overall architecture of the proposed method is illustrated in Fig.~\ref{fig:structure}. In this section, we mainly introduce the co-attention based encoder fusion network and the boundary enhanced decoder network.
\begin{figure*}[t]
\centering
\begin{tabular}{c@{}c}
\includegraphics[width=0.96\linewidth]{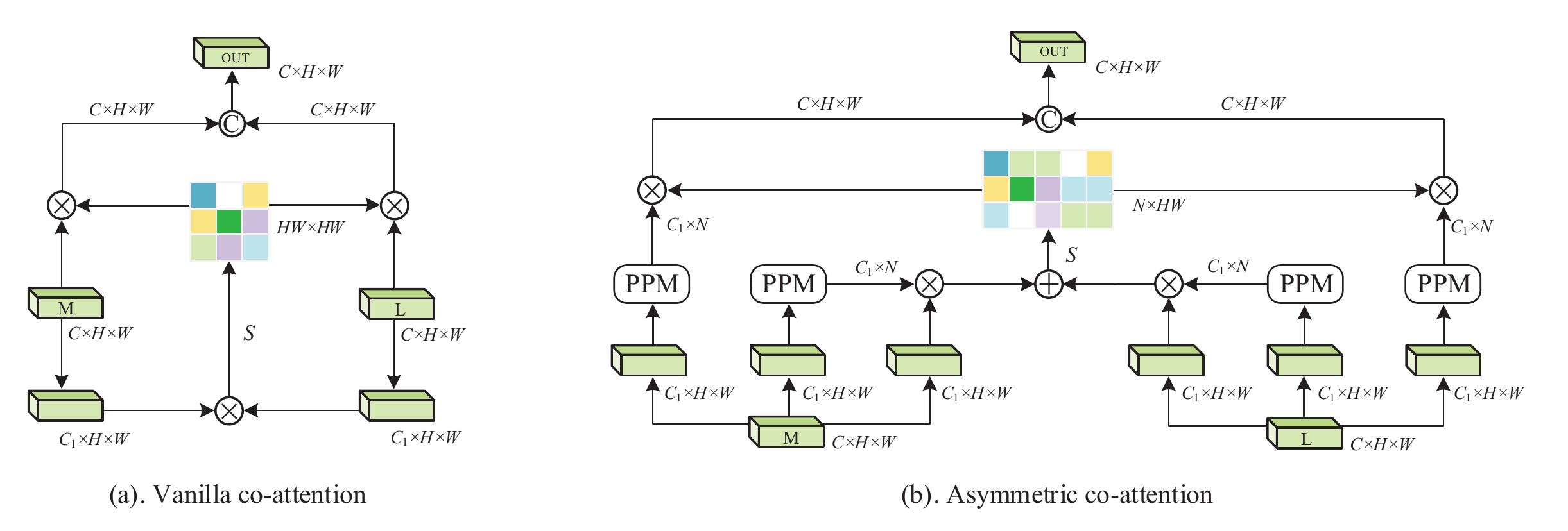}
\ \\
\end{tabular}
{\begin{center}
\vspace{-9mm}
\vspace{0mm}
\caption{
\small{Two co-attention modules. $\mathrm{M}$: Initial multi-modal features. $\mathrm{L}$: Adaptive linguistic context. S: Softmax. PPM: Pyramid pooling module. \textcircled{\fontsize{7pt}{\baselineskip}\selectfont C}: Concatenation.  \textcircled{\fontsize{9pt}{\baselineskip}\selectfont $\times$}: Matrix multiplication. \textcircled{\fontsize{9pt}{\baselineskip}\selectfont $+$}: Element-wise summation. $C$, $H$ and $W$ are channel number, height and width of feature maps, respectively.
} }
\vspace{-1cm}
\label{fig:CAM}
\end{center}
}
\end{figure*}

\subsection{Encoder Fusion with Co-Attention}
\noindent\textbf{Encoder fusion network.}
For a input image, we use ResNet101~\cite{zagoruyko2016wide} to extract visual features. The ResNet101 is composed of five basic blocks: $conv1$, $res2$, $res3$, $res4$, and $res5$. The feature maps from these five blocks are represented as $\{{\mathrm{E}}_i\}^5_{i=1}$. To avoid losing excessive spatial details, the stride of the last block is set to 1. Unlike previous method that performs multi-modal fusion in the decoder, we insert language features after $res3$, $res4$, and $res5$ respectively. ResNet is converted into a multi-modal feature extractor. This design takes full advantage of the data fitting ability of \textbf{deep} CNN model and realizes the deep interleaving of cross-modal features. The experimental comparison between encoder fusion network (EFN) and decoder fusion network (DFN) is implemented, and the results on the UNC dataset are shown in Tab.~\ref{tab:ablation}.

\noindent\textbf{Multi-modal feature representation.}
For a given expression, we feed the word embeddings $\{e_t\}^T_{t=1}$ into the Bi-GRU to generate the linguistic context $\{h_t\}^T_{t=1}$, where $T$ represents the length of the language. Besides, we adopt a simple concatenation strategy to generate the initial multi-modal feature and denote it as:
\begin{equation}
 m_p={w[e_i^p, h_T, s_i^p]},
\label{initial_fusion}
\end{equation}
where $e_i^p$ is a feature vector of $\mathrm{E}_i$ at position $p$. $s_i^p$ represents 8-D spatial coordinates, which follows the design in~\cite{hu2020bi}. $w$ is the learnable parameters. Then we use $m_p$ to calculate the position-specific linguistic context ${l}_{p}$:
\begin{equation}
\begin{split}
& \alpha_{p,t} =  {m_p}^{\top}  \cdot  e_t,  \\
& {l}_{p} = \sum_{t=1}^{T} { {e_t} \cdot {\frac{\exp(\alpha_{p,t})}{\sum_{t=1}^{T} \exp(\alpha_{p,t})} }}.
\end{split}
\end{equation}
${l}_{p}$ treats each word differently. It can suppress the noise in the language expression and highlight the desired region.
Next, the feature maps $\mathrm{M}=[m_p]$ and ${\mathrm{L}}=[l_p]$ go through the co-attention module to achieve the multi-modal fusion.

\noindent\textbf{Vanilla co-attention.} We design a co-attention scheme, which can model the dependencies between multi-modal features and project the multi-modal features to the common feature subspace.
For the convenience of description, the size of $\mathrm{M}$ is defined as ${C\times H\times W}$, where $H$, $W$ and $C$ represent its height, width and channel number, respectively. The feature ${\mathrm{L}}$ has the same dimension as the $\mathrm{M}$. At first, the features $\mathrm{M}$ and $\mathrm{L}$ are flattened into matrix representations with size ${C\times (HW)}$. Their affinity matrix ${\mathrm{A}}\in \mathbb{R}^{HW\times HW}$ is calculated as follows:
\begin{equation}
\begin{split}
& \mathrm{A} =  ({\mathrm{W}_{{m}}}{\mathrm{M}})^\top ({\mathrm{W}_{{l}}}{\mathrm{L}}), \\
\end{split}
\end{equation}
where $\mathrm{W}_m, \mathrm{W}_l\in \mathbb{R}^{C_1\times C}$ are the learnable parameters. The element $a_{i,j} $ of  $\mathrm{A}$ represents the similarity between the $i^{th}$ position of $\mathrm{M}$ and the $j^{th}$ position of $\mathrm{L}$.

Then, we use the softmax function to normalize the similarity matrix as follows:
\begin{equation}
\begin{split}
& \mathrm{A}_1 = {\rm softmax}(\mathrm{A}),\\
& \mathrm{A}_2 = {\rm softmax}(\mathrm{A}^\top),\\
\end{split}
\end{equation}
where $\mathrm{A}_1$ and $\mathrm{A}_2$ are the results of the row-wise and column-wise normalization, respectively. Thus, the feature maps $\mathrm{M}$ and $\mathrm{L}$ can be updated through weighted summing:
\begin{equation}
\begin{split}
& \widetilde{\mathrm{M}} = \mathrm{M}{\mathrm{A}_1}^\top,\\
& \widetilde{\mathrm{L}} = \mathrm{L}{\mathrm{A}_2}^\top.\\
\end{split}
\end{equation}
We concatenate $\widetilde{\mathrm{M}}$ and $\widetilde{\mathrm{L}}$ along channel dimension and follow with a 3$\times$3 convolution to get the multi-modal features $\mathrm{F}\in \mathbb{R}^{C_2\times H\times W}$.
The $\mathrm{F}$ is normalized and added to the encoder feature $\mathrm{E}$. Thus, the embedding of the multi-modal feature in the encoder is finished.
This mechanism can provide extra complementary cues according to the information of the other modality to implement the mutual guidance between these two modalities.
Fig.~\ref{fig:CAM} (a) shows the detailed structure of vanilla co-attention module (VCM).

\noindent\textbf{Asymmetric co-attention.} Furthermore, we propose an asymmetric co-attention module (ACM) to reduce the computational cost. Inspired by~\cite{zhu2019asymmetric}, we employ pyramid pooling module (PPM) to sample the feature maps $\mathrm{M}$ and $\mathrm{L}$. The PPM is composed of four-scale feature bins, which
are then flattened and concatenated to form a matrix of size $C_1$$\times$$N$,  $N \ll HW$.
Here, the sizes of the feature bins are set to 1$\times$1, 3$\times$3, 6$\times$6 and 8$\times$8, respectively.
Thus, the self-affinity matrixes of $\mathrm{M}$ and $\mathrm{L}$ can be calculated as:
\begin{equation}
\begin{split}
& {\rm SA}_m=({\rm PPM}(\mathrm{W}_m^1\mathrm{M}))^\top(\mathrm{W}_m^2\mathrm{M}),\\
& {\rm SA}_l=({\rm PPM}(\mathrm{W}_l^1{\mathrm{L}}))^\top(\mathrm{W}_l^2{\mathrm{L}}),\\
\end{split}
\end{equation}
where ${\rm SA}_{m}$ and ${\rm SA}_l$ denote the modality-specific similarity matrixes. Their sizes are fixed to $N$$\times$$(HW)$ through the PPM, which is asymmetric. $\mathrm{W}_m^1$, $\mathrm{W}_m^2$, $\mathrm{W}_l^1$ and $\mathrm{W}_l^2$ indicate the learnable parameters. We further combine these two matrices as follows:
\begin{equation}
\begin{split}
& {\rm A}_3={\rm softmax}(({\rm SA}_m + {\rm SA}_l)^\top).\\
\end{split}
\end{equation}
Then, the row-wise normalized matrix ${\rm A}_3 \in \mathbb{R}^{(HW)\times N}$ is used to assist the update of multi-modal features:
\begin{equation}
\begin{split}
& \widetilde{\mathrm{M}}={\rm A}_3 ({\rm PPM}(\mathrm{W}_m^3{\mathrm{M}}))^\top,\\
& \widetilde{\mathrm{L}}={\rm A}_3 ({\rm PPM}(\mathrm{W}_l^3{\mathrm{L}}))^\top.\\
\end{split}
\end{equation}
Similarly to the vanilla co-attention, $\widetilde{\mathrm{M}}$ and $\widetilde{\mathrm{L}}$ is concatenated to generate the final multi-modal output. The whole structure of ACM is shown in Fig.~\ref{fig:CAM} (b).
\subsection{Boundary Enhancement Module}
In CNNs, the repeated stride and pooling operations lead to the loss of fine structure information, which may blur the  contour of the predicted region. Previous works~\cite{ye2019cross,chen2019see,hu2020bi,huang2020referring,hui2020linguistic} do not explicitly consider the restoration of details when performing multi-scale fusion in decoder. In this work, we design a boundary enhancement module (BEM), which uses boundary features as a guidance to make the network attend to finer details and realize the progressive refinement of the prediction. Its structure is shown in Fig.~\ref{fig:structure}. Specifically, for the decoder features $\{\mathrm{D}_i\}_{i=1}^{5}$, we first compute the boundary-aware features:
\begin{equation}
\label{Eq:decompose}
\begin{split}
& \mathrm{B}_i = \mathrm{S}_i - {\rm STN}(\mathrm{S}_i),\\
\end{split}
\end{equation}
where ${\rm STN}$ represents a spatial transformer networks~\cite{jaderberg2015spatial}. Here, we utilize it to sample the high-level abstract semantic information from $\mathrm{S}_i$. Thus, the residual $\mathrm{B}_i$ describes the fine structure. The prediction process of the boundary map can be written as:
\begin{equation}
\begin{split}
& \widetilde{\mathrm{B}}_{i-1} = Conv(Cat(\mathrm{B}_i, \mathrm{D}_{i-1})),\\
& {\mathrm{{\mathbf{BM}}}_{i-1}} = Sig(Conv(\widetilde{\mathrm{B}}_{i-1})),\\
\end{split}
\end{equation}
where $Cat(\cdot, \cdot)$ is the concatenation operation along the channel axis. $Conv$ and $Sig$ denote the convolutional layer and sigmoid function, respectively. $\mathrm{{\mathbf{BM}}}_{i-1}$ is supervised by the ground-truth contour of the targeted region.

Next, we exploit boundary feature $\widetilde{\mathrm{B}}_{i-1}$ to refine the segmentation mask as follows:
\begin{equation}
\begin{split}
& \mathrm{S}_{i-1} = Conv(Cat(\widetilde{\mathrm{B}}_{i-1}+{\rm STN}(\mathrm{S}_i), \mathrm{S}_i)),\\
& {\mathrm{\mathbf{SM}}_{i-1}} = Sig(Conv(\mathrm{S}_{i-1})),\\
\end{split}
\end{equation}
where $\mathrm{S}_{i-1}$ actually combines the information of decoder features $\mathrm{D}_i$ and $\mathrm{D}_{i-1}$.
${\mathrm{\mathbf{SM}}}_{i-1}$ denotes the refined mask, which is supervised by the ground-truth segmentation. The ${\mathrm{\mathbf{SM}}}_{1}$ from the last decoder block is taken as the final prediction map, as illustrated in Fig.~\ref{fig:structure}.
\begin{table*}[t]
\setlength{\tabcolsep}{4pt}
\small
\centering
\caption{\small{Quantitative evaluation of different methods on four datasets. -: no data available. DCRF: DenseCRF~\cite{krahenbuhl2011efficient} post-processing.}} \vspace{-3mm}
\renewcommand{\arraystretch}{1.0}
\begin{tabular}{p{3.0cm}<{\centering}||p{1.1cm}<{\centering}|p{1.1cm}<{\centering}|p{1.1cm}<{\centering}|p{1.1cm}<{\centering}
|p{1.1cm}<{\centering}|p{1.1cm}<{\centering}|p{1.1cm}<{\centering}|p{1.1cm}<{\centering}}
\hline
\multirow{2}{*}{*}
&\multicolumn{1}{c|}{ReferIt} & \multicolumn{3}{c|}{UNC}
&\multicolumn{3}{c|}{UNC+} &\multicolumn{1}{c}{G-Ref}\\
\cline{2-9}
& test    &val&testA&testB   &val&testA&testB   &val  \\
\hline \hline
LSTM-CNN$_{16}$~\cite{hu2016segmentation}    &48.03 &-     &-     &-     &-     &-     &-     &28.14\\
RMI+DCRF$_{17}$~\cite{liu2017recurrent}           &58.73 &45.18 &45.69 &45.57 &29.86 &30.48 &29.50 &34.52\\
DMN$_{18}$~\cite{margffoy2018dynamic}        &52.81 &49.78 &54.83 &45.13 &38.88 &44.22 &32.29 &36.76\\
KWA$_{18}$~\cite{shi2018key}                 &59.19 &-     &-     &-     &-     &-     &-     &36.92\\
RRN+DCRF$_{18}$~\cite{li2018referring}            &63.63 &55.33 &57.26 &53.95 &39.75 &42.15 &36.11 &36.45\\
MAttNet$_{18}$~\cite{yu2018mattnet}          &-     &56.51 &62.37 &51.70 &46.67 &52.39 &40.08 &-    \\
lang2seg$_{19}$~\cite{Chen_lang2seg_2019}    &-     &58.90 &61.77 &53.81 &-     &-     &-     &-    \\
CMSA+DCRF$_{19}$~\cite{ye2019cross}               &63.80 &58.32 &60.61 &55.09 &43.76 &47.60 &37.89 &39.98\\
STEP$_{19}$~\cite{chen2019see}               &64.13 &60.04 &63.46 &57.97 &48.19 &52.33 &40.41 &46.40\\
CGAN$_{20}$~\cite{luo2020cascade}            &-     &59.25 &62.37 &53.94 &46.16 &51.37 &38.24 &46.54\\
BRINet+DCRF$_{20}$~\cite{hu2020bi}                &63.46 &61.35 &63.37 &59.57 &48.57 &52.87 &42.13 &48.04\\
LSCM+DCRF$_{20}$~\cite{hui2020linguistic}         &66.57 &61.47 &64.99 &59.55 &49.34 &53.12 &\textbf{43.50} &48.05\\
CMPC+DCRF$_{20}$~\cite{huang2020referring}        &65.53 &61.36 &64.54 &59.64 &49.56 &53.44 &43.23 &49.05\\
\hline \hline
Ours(VCM)                            &66.06 &62.53 &65.36 &59.19 &50.24 &55.04 &41.68 &51.22\\  
Ours(ACM)                            &\textbf{66.70} &\textbf{62.76} &\textbf{65.69} &\textbf{59.67} &\textbf{51.50} &\textbf{55.24} &43.01 &\textbf{51.93}\\

\hline
Ours$_{\rm \scriptscriptstyle VCM}^{\rm \scriptscriptstyle coco}$                &-   &69.27 &70.56 &66.36 &57.46 &61.75 &50.66 &57.51\\
Ours$_{\rm \scriptscriptstyle ACM}^{\rm \scriptscriptstyle coco}$                &-   &68.97 &71.13 &66.95 &57.48 &61.35 &51.97 &57.49\\
\hline
\end{tabular}
\vspace{-0mm}
\label{tab:quantitative}
\end{table*}
\begin{figure*}[htp]
\begin{center}
\begin{tabular}{c@{\hspace{0.8mm}}c@{\hspace{0.8mm}}c@{\hspace{0.8mm}}c@{\hspace{0.8mm}}c@{\hspace{0.8mm}}c@{\hspace{0.8mm}}c}
\includegraphics[width=0.16\linewidth,height=0.10\linewidth]{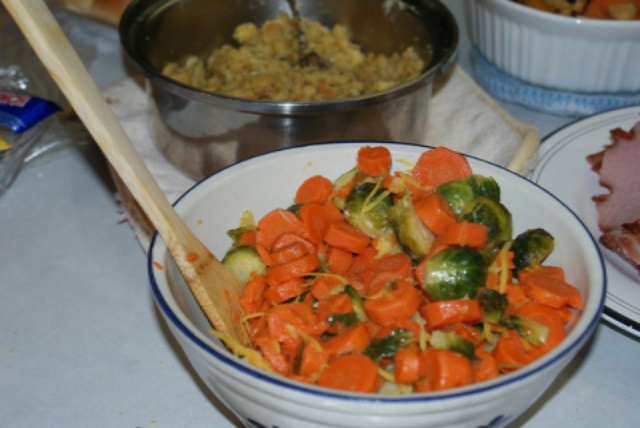}&
\includegraphics[width=0.16\linewidth,height=0.10\linewidth]{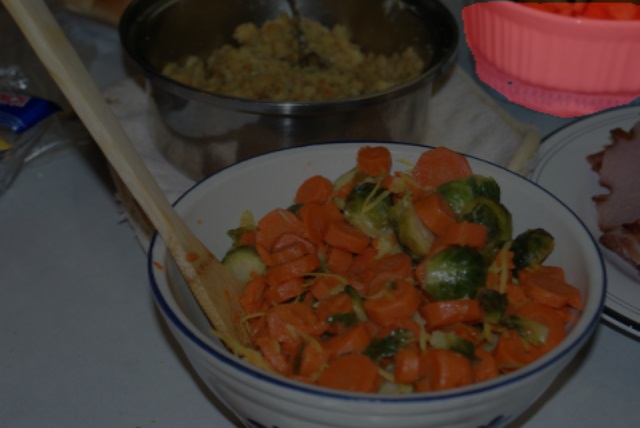}&
\includegraphics[width=0.16\linewidth,height=0.10\linewidth]{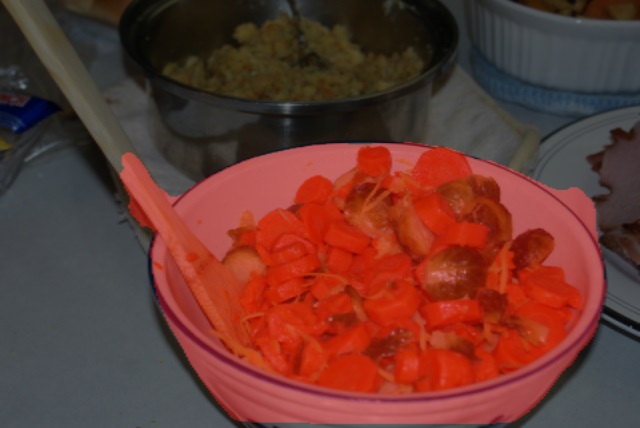}&
\includegraphics[width=0.16\linewidth,height=0.10\linewidth]{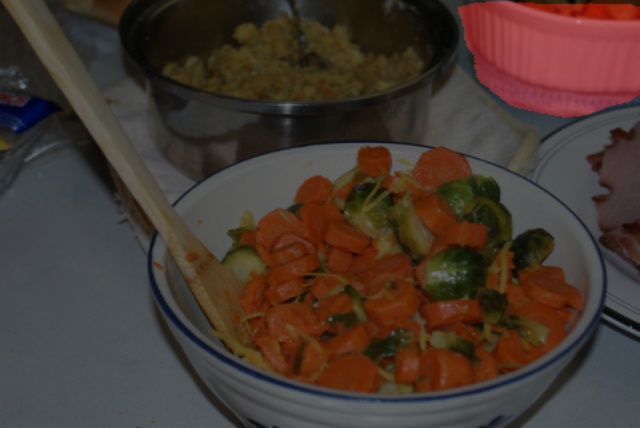}&
\includegraphics[width=0.16\linewidth,height=0.10\linewidth]{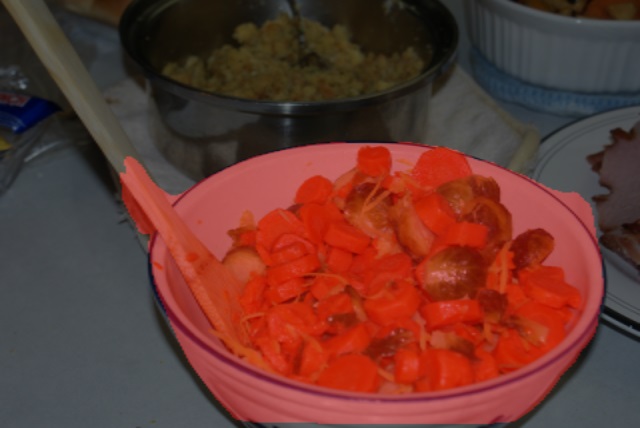}&
\includegraphics[width=0.16\linewidth,height=0.10\linewidth]{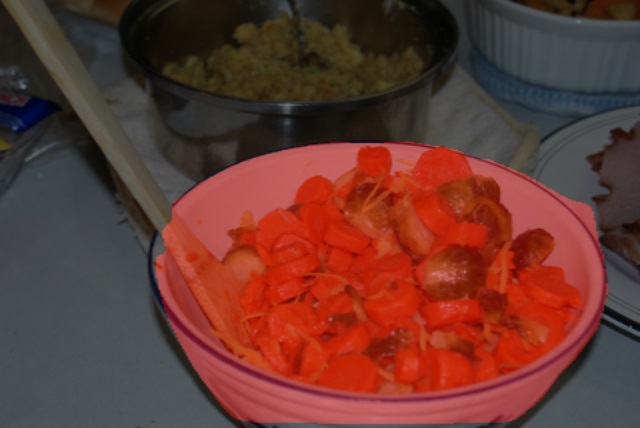}\vspace{-0.7mm}\\
\fontsize{8pt}{\baselineskip}\selectfont Image&
\fontsize{7pt}{\baselineskip}\selectfont  \makecell[c]{dish in top right corner}&
\fontsize{7pt}{\baselineskip}\selectfont  \makecell[c]{bowl of carrots}&
\fontsize{7pt}{\baselineskip}\selectfont  \makecell[c]{white dish in the top \\ right corner}&
\fontsize{7pt}{\baselineskip}\selectfont  \makecell[c]{carrots}&
\fontsize{7pt}{\baselineskip}\selectfont  \makecell[c]{front bowl with carrots in it}\vspace{2mm}\\
\includegraphics[width=0.16\linewidth,height=0.10\linewidth]{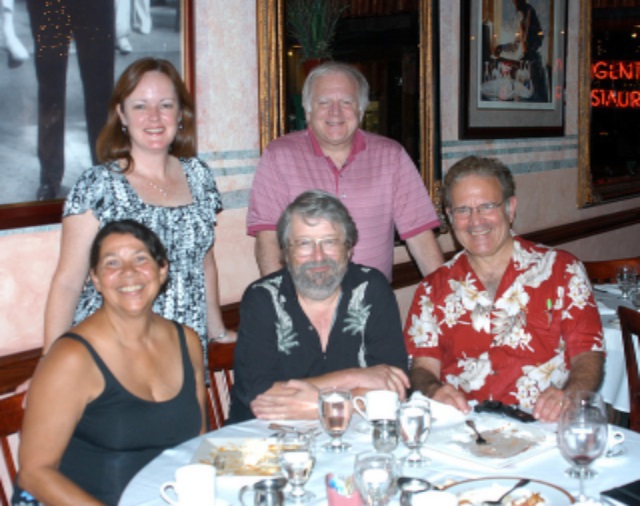}&
\includegraphics[width=0.16\linewidth,height=0.10\linewidth]{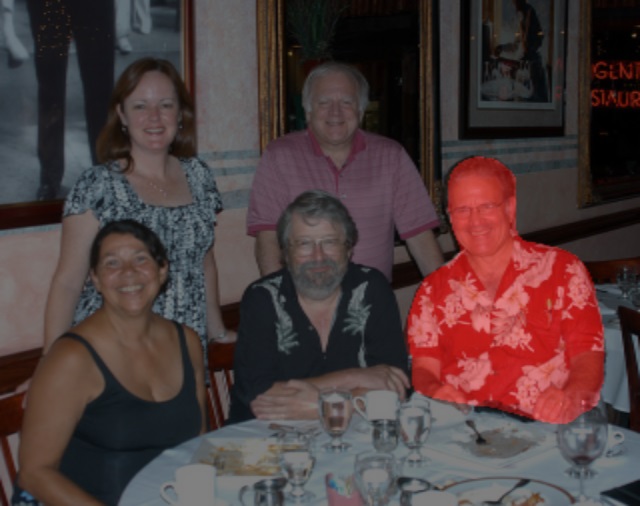}&
\includegraphics[width=0.16\linewidth,height=0.10\linewidth]{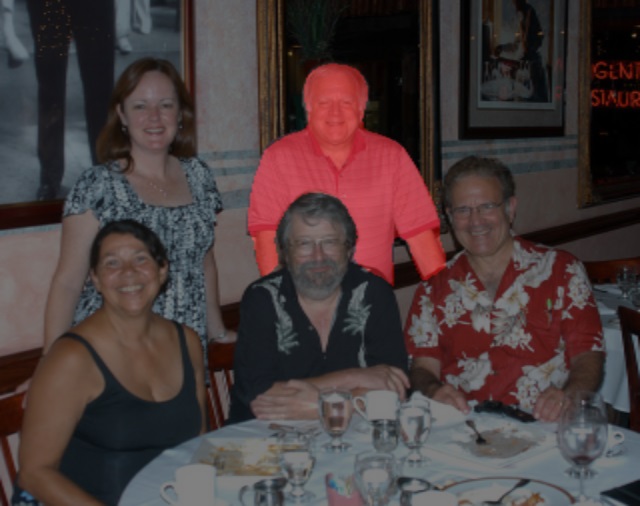}&
\includegraphics[width=0.16\linewidth,height=0.10\linewidth]{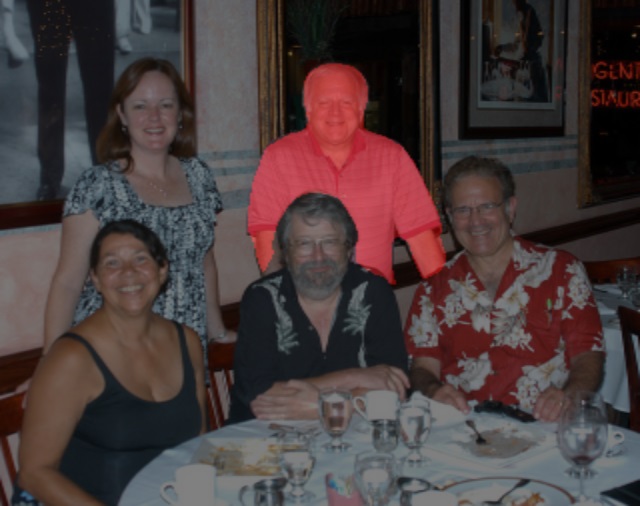}&
\includegraphics[width=0.16\linewidth,height=0.10\linewidth]{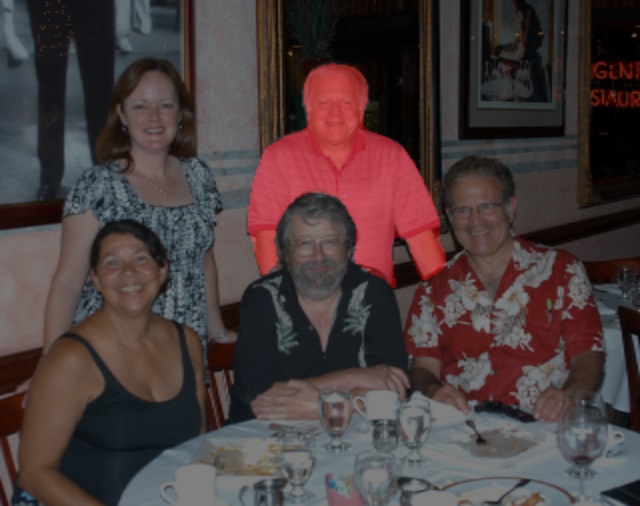}&
\includegraphics[width=0.16\linewidth,height=0.10\linewidth]{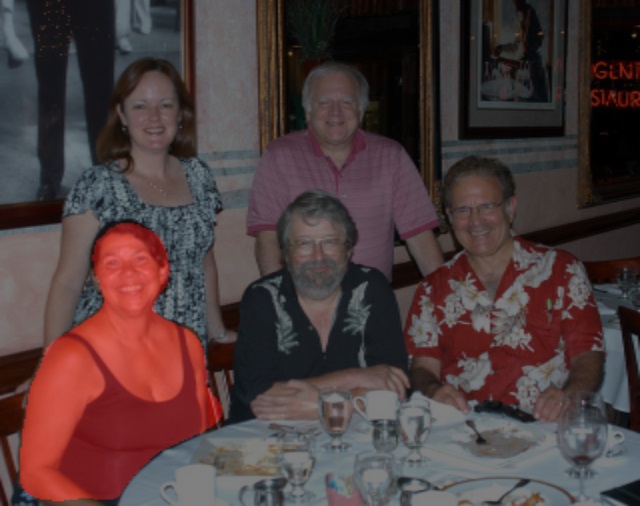}\vspace{-0.7mm}\\
\fontsize{8pt}{\baselineskip}\selectfont Image&
\fontsize{7pt}{\baselineskip}\selectfont  \makecell[c]{red white shirt}&
\fontsize{7pt}{\baselineskip}\selectfont  \makecell[c]{pink shirt top}&
\fontsize{7pt}{\baselineskip}\selectfont  \makecell[c]{pink shirt guy in back}&
\fontsize{7pt}{\baselineskip}\selectfont  \makecell[c]{guy in pink shirt}&
\fontsize{7pt}{\baselineskip}\selectfont  \makecell[c]{woman front row}\vspace{2mm}\\
\includegraphics[width=0.16\linewidth,height=0.10\linewidth]{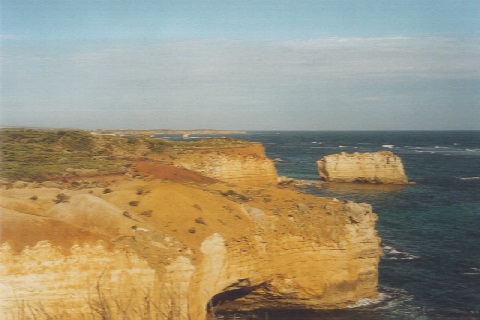}&
\includegraphics[width=0.16\linewidth,height=0.10\linewidth]{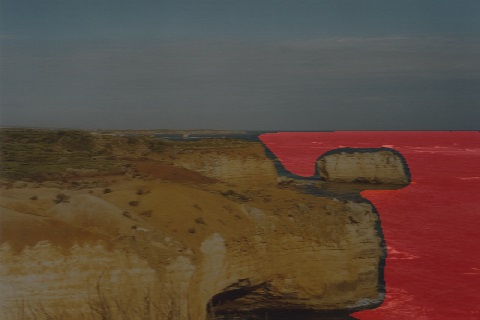}&
\includegraphics[width=0.16\linewidth,height=0.10\linewidth]{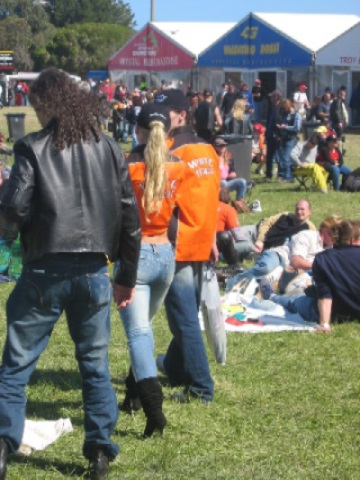}&
\includegraphics[width=0.16\linewidth,height=0.10\linewidth]{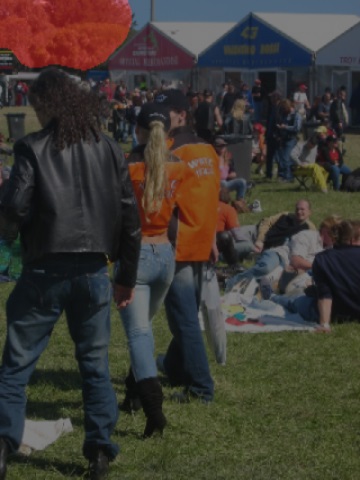}&
\includegraphics[width=0.16\linewidth,height=0.10\linewidth]{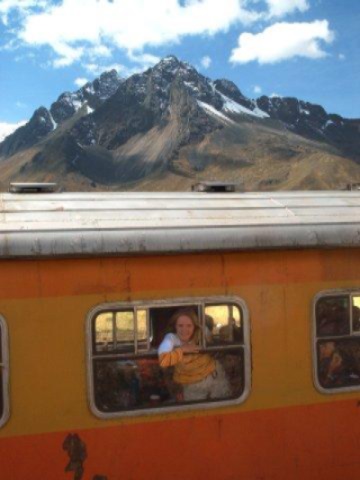}&
\includegraphics[width=0.16\linewidth,height=0.10\linewidth]{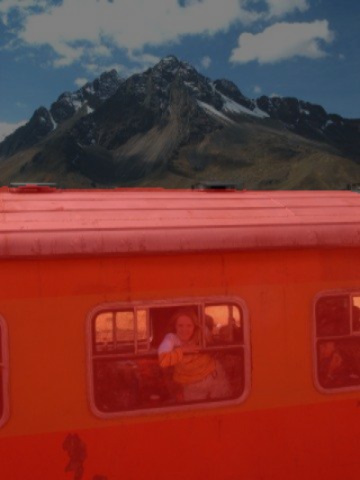}\vspace{-0.7mm}\\
\fontsize{8pt}{\baselineskip}\selectfont Image&
\fontsize{7pt}{\baselineskip}\selectfont  \makecell[c]{water}&
\fontsize{8pt}{\baselineskip}\selectfont Image&
\fontsize{7pt}{\baselineskip}\selectfont  \makecell[c]{trees top left corner}&
\fontsize{8pt}{\baselineskip}\selectfont Image&
\fontsize{7pt}{\baselineskip}\selectfont  \makecell[c]{bottom half of the picture}\vspace{2mm}\\
\end{tabular}
\end{center}
\vspace{-8mm}
\caption{\small{Visual examples of referring image segmentation by our method.}}\label{fig:performance}
\vspace{-1mm}
\end{figure*}
\begin{table}[t]
\vspace{0mm}
\setlength{\tabcolsep}{4pt}
\small
\centering
\caption{IoU for different length referring expressions on Google-Ref, UNC, UNC+ and ReferItGame.} \vspace{-3mm}
\renewcommand{\arraystretch}{1.0}
\begin{tabular}{p{1.1cm}<{\centering}|p{1.8cm}<{\centering}||p{0.8cm}<{\centering}|p{0.8cm}<{\centering}|p{0.8cm}<{\centering}|p{0.8cm}<{\centering}}
\hline
\multirow{1}{*}{} &Length &1-5 &6-7 &8-10 &11-20 \\
\hline \hline
\multirow{5}{*}{G-Ref}
&R+LSTM~\cite{liu2017recurrent}    &32.29 &28.27 &27.33 &26.61 \\
&R+RMI~\cite{liu2017recurrent}     &35.34 &31.76 &30.66 &30.56 \\
&BRINet~\cite{hu2020bi}            &51.93 &47.55 &46.33 &46.49 \\
&Ours(VCM)                         &57.96 &52.19 &48.78 &46.67\\
&Ours(ACM)                         &59.92 &52.94 &49.56 &46.21\\
\hline
\end{tabular}
\vspace{6mm}

\begin{tabular}{p{1.1cm}<{\centering}|p{1.8cm}<{\centering}||p{0.8cm}<{\centering}|p{0.8cm}<{\centering}|p{0.8cm}<{\centering}|p{0.8cm}<{\centering}}
\hline
\multirow{1}{*}{} &Length &1-2 &3 &4-5 &6-20 \\
\hline \hline
\multirow{5}{*}{UNC}
&R+LSTM~\cite{liu2017recurrent}    &43.66 &40.60 &33.98 &24.91 \\
&R+RMI~\cite{liu2017recurrent}     &44.51 &41.86 &35.05 &25.95 \\
&BRINet~\cite{hu2020bi}            &65.99 &64.83 &56.97 &45.65 \\
&Ours(VCM)                         &68.18 &66.14 &56.82 &46.01 \\
&Ours(ACM)                         &68.73 &65.58 &57.32 &45.90 \\
\hline
\end{tabular}
\vspace{6mm}

\begin{tabular}{p{1.1cm}<{\centering}|p{1.8cm}<{\centering}||p{0.8cm}<{\centering}|p{0.8cm}<{\centering}|p{0.8cm}<{\centering}|p{0.8cm}<{\centering}}
\hline
\multirow{1}{*}{} &Length &1-2 &3 &4-5 &6-20 \\
\hline \hline
\multirow{5}{*}{UNC+}
&R+LSTM~\cite{liu2017recurrent}    &34.40 &24.04 &19.31 &12.30 \\
&R+RMI~\cite{liu2017recurrent}     &35.72 &25.41 &21.73 &14.37 \\
&BRINet~\cite{hu2020bi}            &59.12 &46.89 &40.57 &31.32 \\
&Ours(VCM)                         &60.87 &48.88 &43.79 &29.45 \\
&Ours(ACM)                         &61.62 &52.18 &43.46 &31.52\\
\hline
\end{tabular}
\vspace{6mm}

\begin{tabular}{p{1.1cm}<{\centering}|p{1.8cm}<{\centering}||p{0.8cm}<{\centering}|p{0.8cm}<{\centering}|p{0.8cm}<{\centering}|p{0.8cm}<{\centering}}
\hline
\multirow{1}{*}{} &Length &1 &2 &3-4 &5-20 \\
\hline \hline
\multirow{5}{*}{ReferIt}
&R+LSTM~\cite{liu2017recurrent}    &67.64 &52.26 &44.87 &33.81 \\
&R+RMI~\cite{liu2017recurrent}     &68.11 &52.73 &45.69 &34.53 \\
&BRINet~\cite{hu2020bi}            &75.28 &62.62 &56.14 &44.40 \\
&Ours(VCM)                         &77.73 &66.02 &59.74 &45.75 \\
&Ours(ACM)                         &78.19 &66.63 &60.30 &46.18 \\
\hline
\end{tabular}
\vspace{0mm}
\label{tab:word_length}
\vspace{0mm}
\end{table}
\section{Experiments}
\begin{table*}[t]
\setlength{\tabcolsep}{4pt}
\small
\centering
\caption{\small{Ablation study on the UNC val, testA and testB datasets.}} \vspace{-3mm}
\renewcommand{\arraystretch}{1.0}
\begin{tabular}{c|ccccc||c|c|c|c|c|c}
\hline
\multirow{1}{*}{} & DFN &EFN &VCM &ACM &BEM &prec@0.5 &prec@0.6 &prec@0.7 &prec@0.8 &prec@0.9 &overall IoU \\

\hline \hline
\multirow{6}{*}{val}
& \checkmark &           &           &           &           &57.30 &50.40 &42.00 &28.42 &9.00  &52.86\\
&            &\checkmark &           &           &           &64.16 &58.45 &51.16 &37.53 &13.39 &55.87\\
&            &\checkmark &\checkmark &           &           &68.61 &63.02 &54.55 &40.20 &13.67 &59.65\\
&            &\checkmark &           &\checkmark &           &69.22 &64.11 &56.33 &41.67 &15.32 &60.09\\
&            &\checkmark &\checkmark &           &\checkmark &74.07 &68.84 &61.76 &48.74 &20.06 &62.53\\
&            &\checkmark &           &\checkmark &\checkmark &73.95 &69.58 &62.59 &49.61 &20.63 &62.76\\
\hline \hline
\multirow{6}{*}{testA}
& \checkmark &           &           &           &           &61.66 &54.30 &44.55 &31.09 &9.32  &55.98\\
&            &\checkmark &           &           &           &67.03 &61.59 &53.92 &40.13 &13.31 &58.07\\
&            &\checkmark &\checkmark &           &           &72.14 &66.80 &58.21 &43.03 &13.19 &62.10\\
&            &\checkmark &           &\checkmark &           &72.95 &67.90 &59.98 &45.04 &14.46 &62.46\\
&            &\checkmark &\checkmark &           &\checkmark &77.53 &73.18 &66.02 &52.11 &18.88 &65.36\\
&            &\checkmark &           &\checkmark &\checkmark &77.66 &73.73 &66.70 &52.75 &19.66 &65.69\\
\hline \hline
\multirow{6}{*}{testB}
& \checkmark &           &           &           &           &52.31 &45.48 &37.51 &26.85 &10.40 &49.66\\
&            &\checkmark &           &           &           &58.61 &52.62 &45.67 &34.56 &15.03 &52.48\\
&            &\checkmark &\checkmark &           &           &64.53 &57.96 &50.54 &37.94 &16.39 &56.76\\
&            &\checkmark &           &\checkmark &           &65.02 &58.33 &50.34 &38.74 &16.31 &57.09\\
&            &\checkmark &\checkmark &           &\checkmark &69.74 &63.75 &56.90 &45.20 &22.51 &59.19\\
&            &\checkmark &           &\checkmark &\checkmark &69.66 &65.14 &58.31 &46.18 &22.43 &59.67\\
\hline \hline
\end{tabular}
\label{tab:ablation}
\end{table*}
\begin{figure*}[t]
\vspace{-2mm}
\begin{center}
\begin{tabular}{c@{\hspace{0.5mm}}c@{\hspace{0.5mm}}c@{\hspace{0.5mm}}c@{\hspace{0.5mm}}c@{\hspace{0.5mm}}c@{\hspace{0.5mm}}c}
\multicolumn{6}{c}{ \rule{0pt}{10pt} {Query:  $``tall \ suitcase"$ }}\\
\includegraphics[width=0.16\linewidth,height=0.10\linewidth]{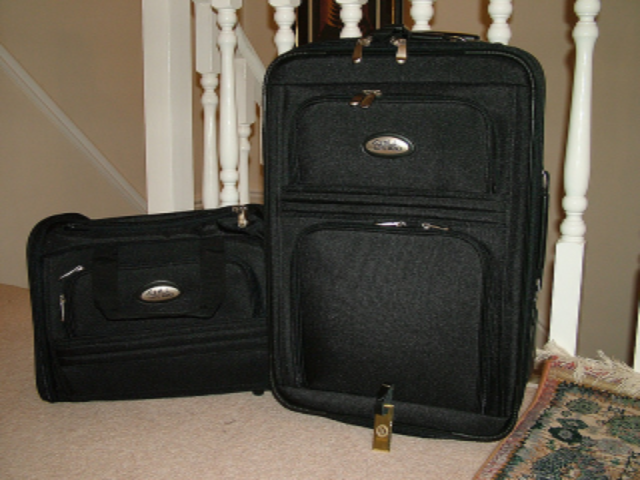}&
\includegraphics[width=0.16\linewidth,height=0.10\linewidth]{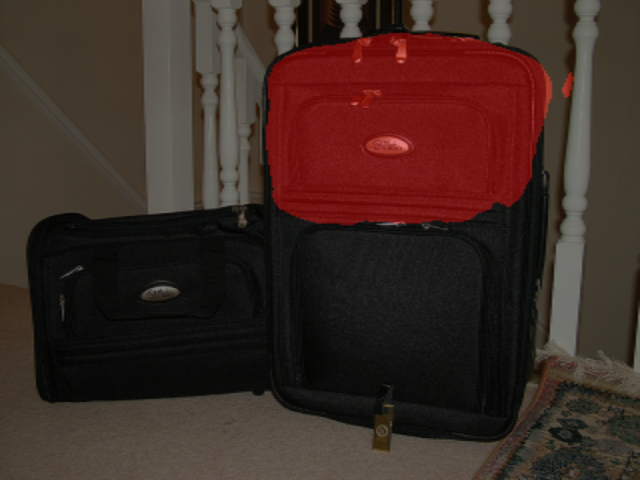}&
\includegraphics[width=0.16\linewidth,height=0.10\linewidth]{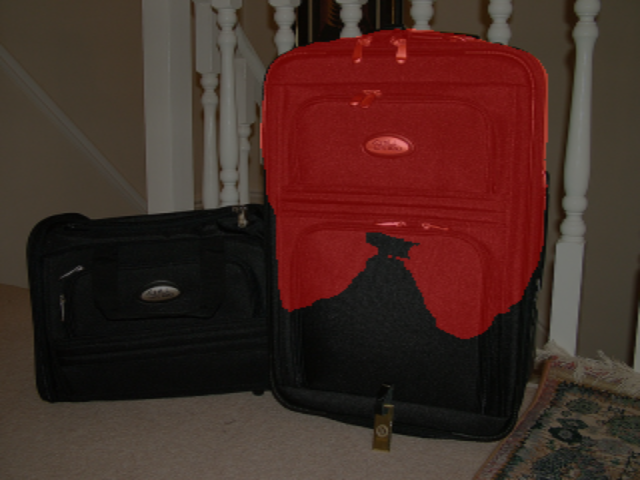}&
\includegraphics[width=0.16\linewidth,height=0.10\linewidth]{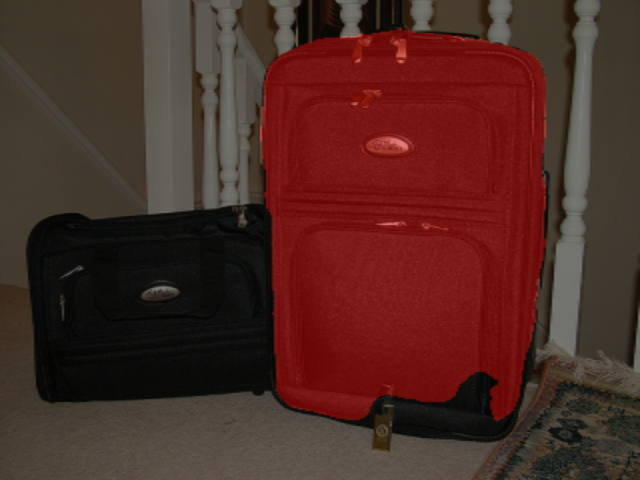}&
\includegraphics[width=0.16\linewidth,height=0.10\linewidth]{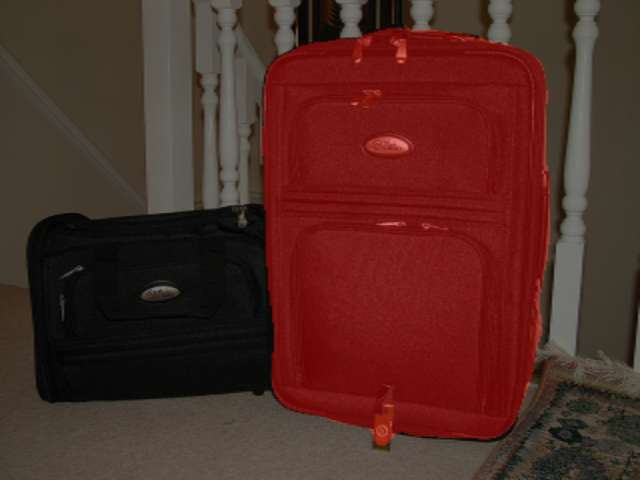}&
\includegraphics[width=0.16\linewidth,height=0.10\linewidth]{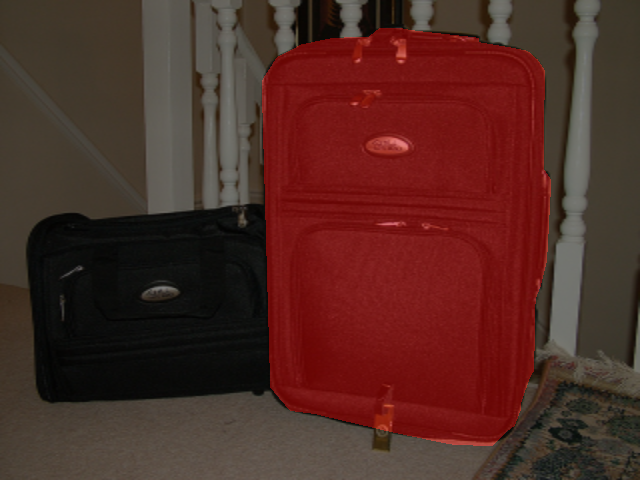}&\vspace{2mm}\\
\multicolumn{6}{c}{ \rule{0pt}{10pt} {Query:  $``chair \ on \ right \ man \ with \ white \ shirt \ sitting \ in \ it"$ }}\\
\includegraphics[width=0.16\linewidth,height=0.10\linewidth]{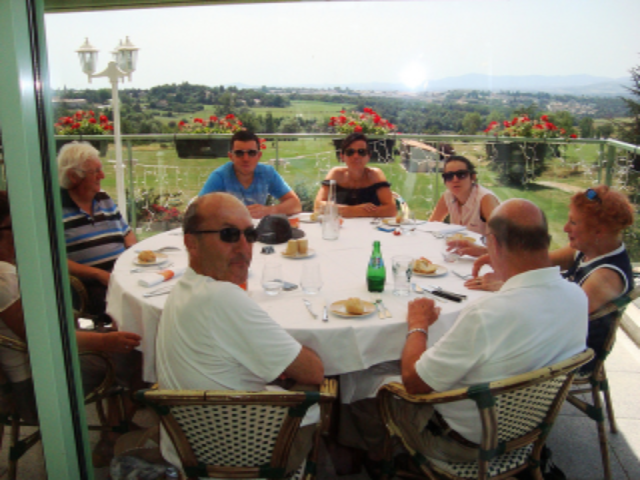}&
\includegraphics[width=0.16\linewidth,height=0.10\linewidth]{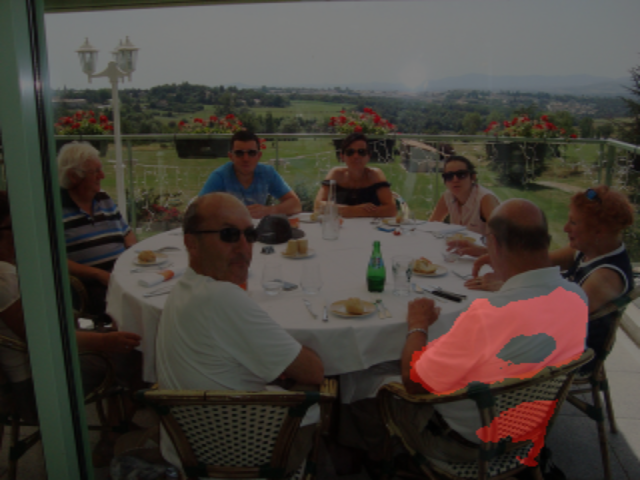}&
\includegraphics[width=0.16\linewidth,height=0.10\linewidth]{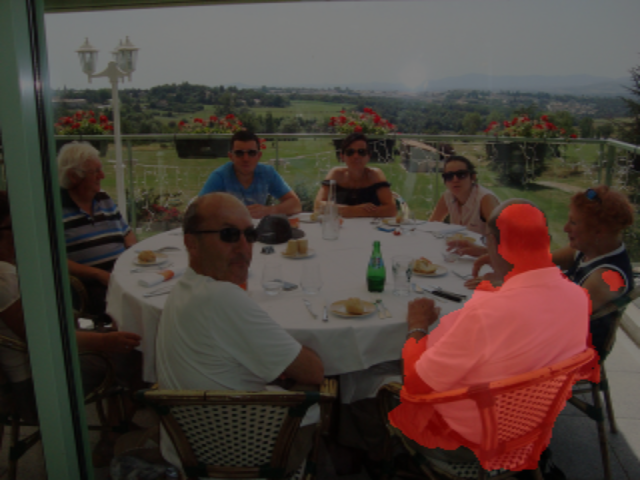}&
\includegraphics[width=0.16\linewidth,height=0.10\linewidth]{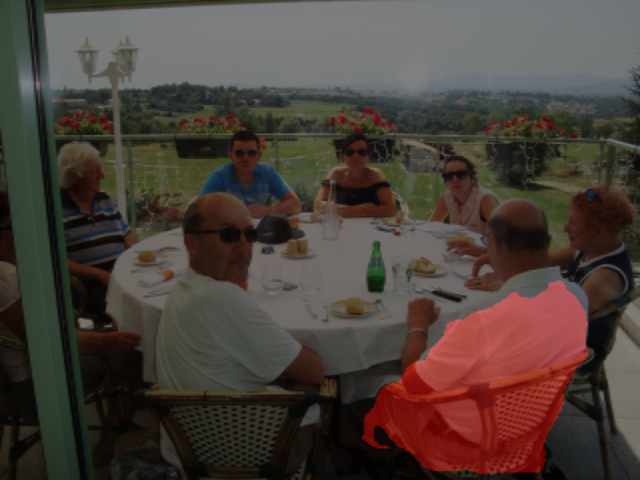}&
\includegraphics[width=0.16\linewidth,height=0.10\linewidth]{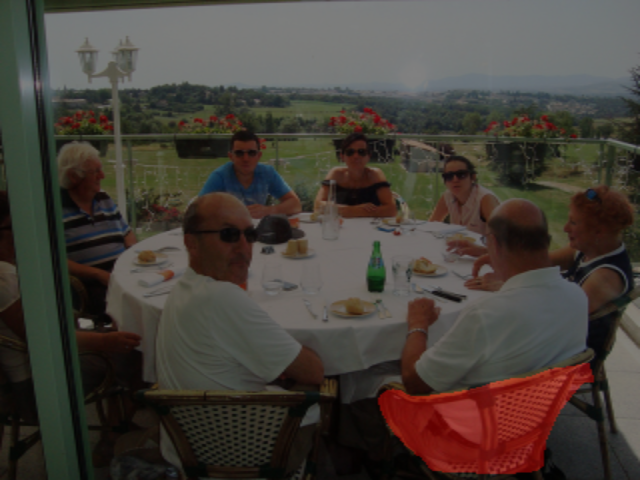}&
\includegraphics[width=0.16\linewidth,height=0.10\linewidth]{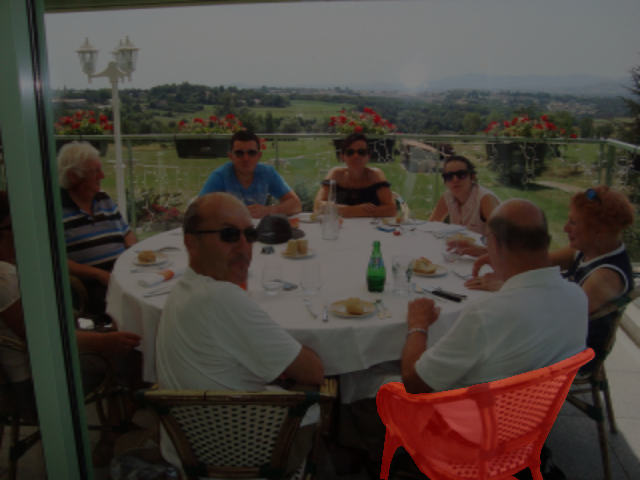}&\vspace{2mm}\\
\fontsize{8.0pt}{\baselineskip}\selectfont Image&
\fontsize{8.0pt}{\baselineskip}\selectfont DFN&
\fontsize{8.0pt}{\baselineskip}\selectfont EFN&
\fontsize{8.0pt}{\baselineskip}\selectfont EFN+ACM&
\fontsize{8.0pt}{\baselineskip}\selectfont EFN+ACM+BEM&
\fontsize{8.0pt}{\baselineskip}\selectfont GT\\
\end{tabular}
\end{center}
\vspace{-5mm}
\caption{Visual examples of the proposed modules.}\label{fig:ablation_1}
\vspace{-1mm}
\end{figure*}
\subsection{Datasets}
To verify the effectiveness of the proposed method, we evaluate the performance on four datasets, which are the UNC~\cite{yu2016modeling}, UNC+~\cite{yu2016modeling}, Google-Ref~\cite{mao2016generation} and ReferIt~\cite{kazemzadeh2014referitgame}.

\textbf{UNC}: It contains 19,994 images with 142,209 language expressions for 50,000 segmented object regions. These data are selected from the MS COCO dataset using a two-player game~\cite{kazemzadeh2014referitgame}. There are multiple objects with the same category in each image.

\textbf{UNC+}: It is also a subset of the MS COCO, which contains 141,564 language expressions for 49,856 objects in 19,992 images. However, the referring expression does not contain the words that indicate location information, which means that the matching of their language and visual region totally depend on the appearance information.

\textbf{Google-Ref}: It includes 104,560 referring expressions for 54,822 objects in 26,711 images.
The annotations are based on Mechanical Turk instead of using a two-player game. The average length of referring expressions in this dataset is 8.43 words.

\textbf{ReferIt}: It is collected from the IAPR TC-12~\cite{escalante2010segmented}. It is composed of 130,525 referring expressions for 96,654 object regions in 19,894 natural images. In addition, their annotations contain objects or stuff, and the expressions are usually shorter and more succinct than the other datasets.
\begin{table*}[t]
\setlength{\tabcolsep}{4pt}
\small
\centering
\caption{Runtime analysis of different methods. The time of post-processing is ignored.} \vspace{-3mm}
\renewcommand{\arraystretch}{1.0}
\begin{tabular}{c||p{1.5cm}<{\centering}|p{1.5cm}<{\centering}|p{1.5cm}<{\centering}|p{1.5cm}<{\centering}|p{1.5cm}<{\centering}|p{1.5cm}<{\centering}|p{1.5cm}<{\centering}|p{1.5cm}<{\centering}}
\hline
\multirow{1}{*}{}          &LSTM &RMI  &RRN  &CMSA &BRINet &CMPC &Ours(VCM) &Ours(ACM)\\
\hline \hline
\multirow{1}{*}{Time(ms)}  &58ms &72ms &43ms &79ms &117ms  &60ms &17ms      &20ms                 \\
\hline
\end{tabular}
\vspace{0mm}
\label{tab:runtime}
\vspace{-3mm}
\end{table*}
\subsection{Implementation Details}
The proposed framework is built on the public pytorch toolbox and is trained on an Nvidia GTX 1080Ti GPU for 200,000 iterations. Our network is trained by an end-to-end strategy and using the SGD optimizer with an initial learning rate of 0.00075 and divided by 10 after 100,000 iterations. All input images are resized to 320$\times$320. The weight decay and batch size are 0.0005 and 12, respectively.
And when training G-ref, we use the UNC model as a pre-training model to avoid over-fitting.
During the inference phase, the prediction map is resized to the same resolution as the original image.
The binary cross entropy loss is used to supervise the boundary map and segmentation map, In addition, we also use the ground-truth segmentation (GT) to supervise the output of the STN.

\textbf{Evaluation Metrics}: Following previous works~\cite{hu2020bi,huang2020referring,hui2020linguistic}, we employ Overall Intersection-over-Union (Overall IoU) and Prec@X to evaluate the segmentation accuracy. The Overall IoU metric represents the ratio of the total intersection regions and the total union regions between the predicted mask and the ground truth for all the test samples. The Prec@X metric calculates the percentage of the IoU score of the prediction mask in the test set that exceeds the threshold $\rm X$, where $\rm X \in \{0.5, 0.6, 0.7, 0.8, 0.9\}$.
\subsection{Performance Comparison}
To verify the effectiveness of the proposed model, we compare it with thirteen methods, which are the
LSTM-CNN~\cite{hu2016segmentation},
RMI~\cite{liu2017recurrent},
DMN~\cite{margffoy2018dynamic},
KWA~\cite{shi2018key},
RRN~\cite{li2018referring},
MAttNet~\cite{yu2018mattnet},
lang2seg~\cite{Chen_lang2seg_2019},
CMSA~\cite{ye2019cross},
STEP~\cite{chen2019see},
CGAN~\cite{luo2020cascade},
BRINet~\cite{hu2020bi},
LSCM~\cite{hui2020linguistic},
and CMPC~\cite{huang2020referring}.

\textbf{Performance Evaluation}: Tab.~\ref{tab:quantitative} shows the performance (IoU) comparison of different methods on four datasets, in which Our(VCM) and Our(ACM) represent the results of using vanilla co-attention module and asymmetric co-attention module, respectively. The proposed model consistently outperforms these competitors on most datasets except the UNC+ testB.
Some methods like LSCM and CMPC apply DenseCRF~\cite{krahenbuhl2011efficient} to refine their final masks while our model does not need any post-processing.
In particular, we achieve the gain of 5.9\%, 3.4\% and 3.9\% over the second best method CMPC~\cite{huang2020referring} on the G-Ref, UNC+ testA and val, respectively.
In addition,
because the UNC, UNC+ and G-Ref are all collected from the MS COCO dataset, we combine their training data into a larger training set. The results of the model trained on it are denoted as Ours$_{\rm \scriptscriptstyle VCM}^{\rm \scriptscriptstyle coco}$ and Ours$_{\rm \scriptscriptstyle ACM}^{\rm \scriptscriptstyle coco}$, which show that sufficient training data can yield better results.
We give some visual examples in Fig.~\ref{fig:performance}. It can be seen that our method can accurately segment the specific regions (object or stuff) according to the query expression. Following~\cite{liu2017recurrent,hu2020bi}, we analyze the relationship between language length and segmentation accuracy. The results are demonstrated in Tab.~\ref{tab:word_length}, which indicate that our method achieves the state-of-the-art performance.

\textbf{Runtime and Memory Statistics}: We implement all the tests on a NVIDIA GTX 1080 Ti GPU. The comparison of running time is reported in Tab.~\ref{tab:runtime}.
Our method runs the fastest with a speed of 50 FPS.
The GPU memory usage is shown in Tab.~\ref{tab:gpumemory}. From Tab.~\ref{tab:runtime} and Tab.~\ref{tab:gpumemory}, we can find that although the VCM has advantages in speed, the large input size causes the memory usage to sharply increase. On the contrary, the VCM is not sensitive to the input size. Therefore, it is widely applicable.
\begin{table}[t]
\setlength{\tabcolsep}{4pt}
\small
\centering
\caption{GPU memory (MB) comparisons between VCM and ACM. The lower values are the better.} \vspace{-3mm}
\renewcommand{\arraystretch}{1.0}
\begin{tabular}{p{1.5cm}<{\centering}||c|c|c}
\hline
\multirow{1}{*}{Input size}          &512$\times$20$\times$20    &512$\times$40$\times$40 &512$\times$96$\times$96\\
\hline \hline
\multirow{1}{*}{VCM} &9.93                       &54.35                   &1308.00                  \\
\multirow{1}{*}{ACM} &6.92                       &14.29                   &61.11                 \\
\hline
\end{tabular}
\vspace{0mm}
\label{tab:gpumemory}
\vspace{-2mm}
\end{table}
\subsection{Ablation Study}
we conduct a series of experiments on the UNC dataset to verify the benefit of each component.

\textbf{Comparison of DFN and EFN}: We first remove the co-attention module and boundary enhancement module from the CEFNet in Fig.~\ref{fig:structure}. Then, we achieve the multi-modal fusion in the encoder by Eq. (\ref{initial_fusion}), and the decoder adopts the FPN~\cite{lin2017feature} structure. This network is taken as the baseline network (EFN) of encoder fusion.
In addition, similar to previous works, we realize the multi-modal fusion in the FPN decoder by Eq. (\ref{initial_fusion}) to build the baseline (DFN) for decoder fusion. We evaluate the two baselines in Tab.~\ref{tab:ablation}, from which we can see that EFN is significantly better than DFN. With the help of ResNet, the encoder fusion strategy achieves more powerful feature coding without increasing additional computational burden.

\textbf{Effectiveness of Co-Attention}: We evaluate the performance of the vanilla co-attention module (VCM) and asymmetric co-attention module(ACM). Compared with the baseline EFN, the VCM brings 6.8\%, 6.9\% and 8.2\% IoU improvement on the UNC-val, UNC-testA, and UNC-testB, respectively. Similarly, the ACM achieves the gain of 7.6\%, 7.6\% and 8.8\% on the same datasets, respectively. The ACM performs slightly better than the VCM. We attribute it to the modality-specific affinity learning, which focuses on important regions within the modality and achieves better contextual understanding of the modality itself. It is conducive to cross-modal alignment in the next stage.

\textbf{Effectiveness of BEM}: Tab.~\ref{tab:ablation} presents the ablation results of boundary enhancement module (BEM), which show that the special consideration of boundary refinement can significantly improve the performance.
BEM can bring about 2\%$\sim$3\% performance improvement (Overall IoU)) to the final prediction result.
Some visual results in Fig.~\ref{fig:ablation_1} demonstrate the benefits of BEM. These figures show that the prediction mask can fit the object boundary more closely after the refinement of BEM.
\subsection{Failure Cases}
\begin{figure}[t]
\vspace{-1mm}
\begin{center}
\begin{tabular}{c@{\hspace{0.5mm}}c@{\hspace{0.5mm}}c@{\hspace{0.8mm}}c@{\hspace{0.5mm}}c@{\hspace{0.5mm}}c@{\hspace{0.5mm}}c}
\multicolumn{3}{c}{  {\fontsize{6.0pt}{\baselineskip}\selectfont { Query:  $``\rm second \ guy \ roght"$ }}}&
\multicolumn{3}{c}{  {\fontsize{6.0pt}{\baselineskip}\selectfont { Query:  $``\rm left \ brown \ horse \ butt"$ }}} \vspace{-0.6mm} \\
\includegraphics[width=0.15\linewidth,height=0.10\linewidth]{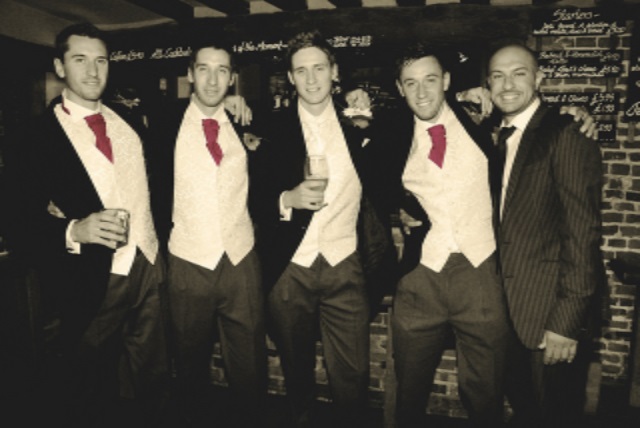}&
\includegraphics[width=0.15\linewidth,height=0.10\linewidth]{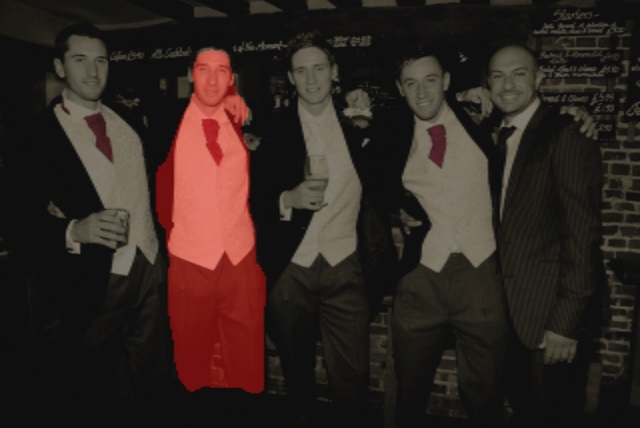}&
\includegraphics[width=0.15\linewidth,height=0.10\linewidth]{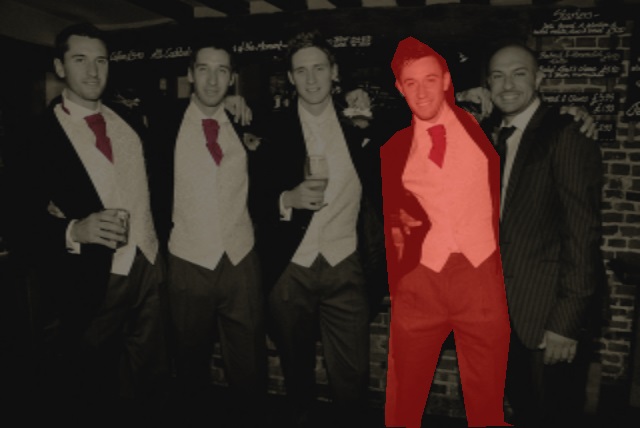}&
\includegraphics[width=0.15\linewidth,height=0.10\linewidth]{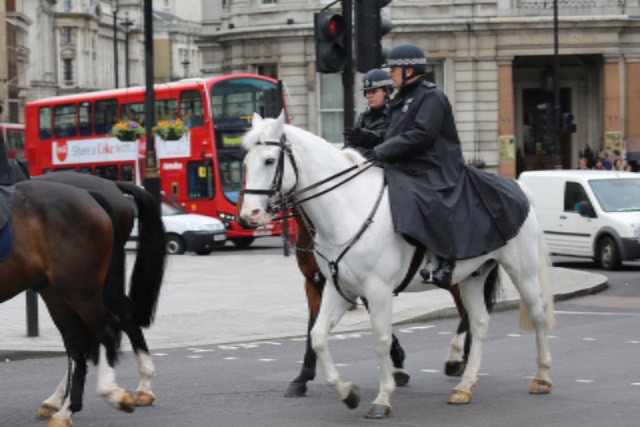}&
\includegraphics[width=0.15\linewidth,height=0.10\linewidth]{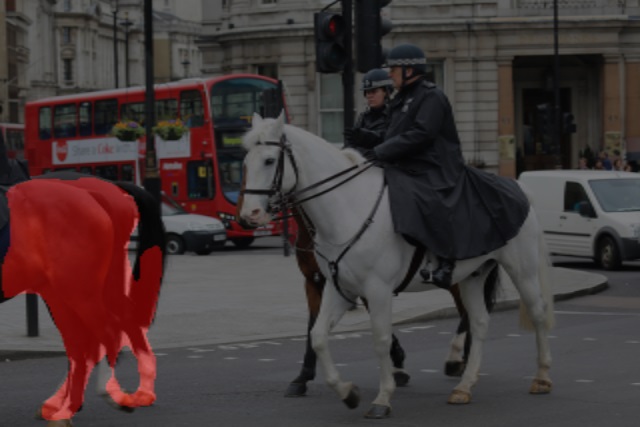}&
\includegraphics[width=0.15\linewidth,height=0.10\linewidth]{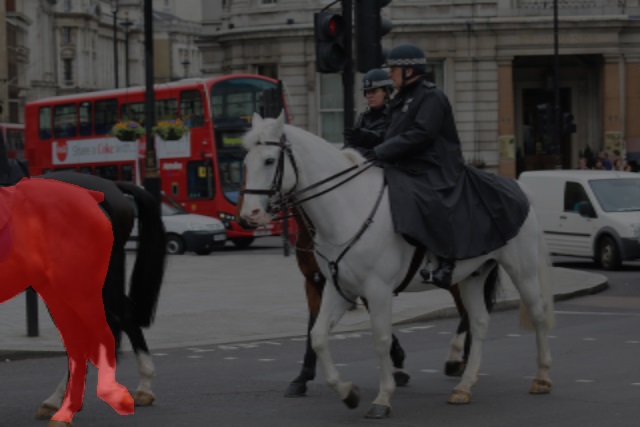}&\vspace{-2.6mm}\\
\fontsize{5pt}{\baselineskip}\selectfont Image&
\fontsize{5pt}{\baselineskip}\selectfont  \makecell[c]{Result}&
\fontsize{5pt}{\baselineskip}\selectfont  \makecell[c]{GT}&
\fontsize{5pt}{\baselineskip}\selectfont Image&
\fontsize{5pt}{\baselineskip}\selectfont  \makecell[c]{Result}&
\fontsize{5pt}{\baselineskip}\selectfont  \makecell[c]{GT}&\vspace{-0.6mm}\\
\multicolumn{3}{c}{ }&
\multicolumn{3}{c}{  {\fontsize{6.0pt}{\baselineskip}\selectfont { Query:  $``\rm the \ grass \ to \ the \ left \ of \ $ }}}& \vspace{-1.6mm} \\
\multicolumn{3}{c}{ {\fontsize{6.0pt}{\baselineskip}\selectfont { Query:  $``\rm middle \ cop"$ }}}&
\multicolumn{3}{c}{  {\fontsize{6.0pt}{\baselineskip}\selectfont { $\rm the \ red \ shirt \ green \ pant \ man"$ }}}  \vspace{-0.6mm}\\
\includegraphics[width=0.15\linewidth,height=0.10\linewidth]{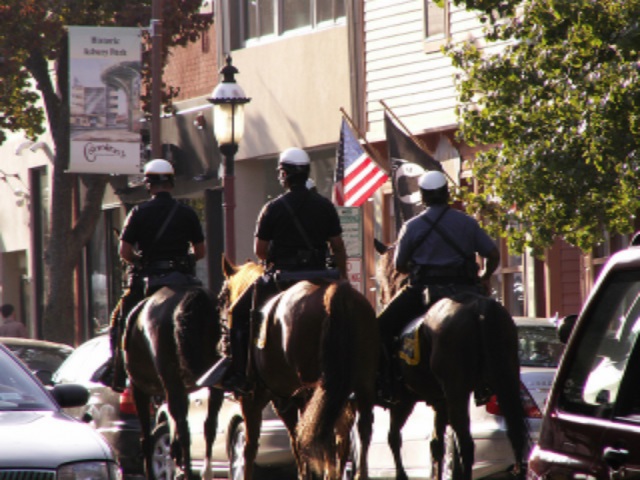}&
\includegraphics[width=0.15\linewidth,height=0.10\linewidth]{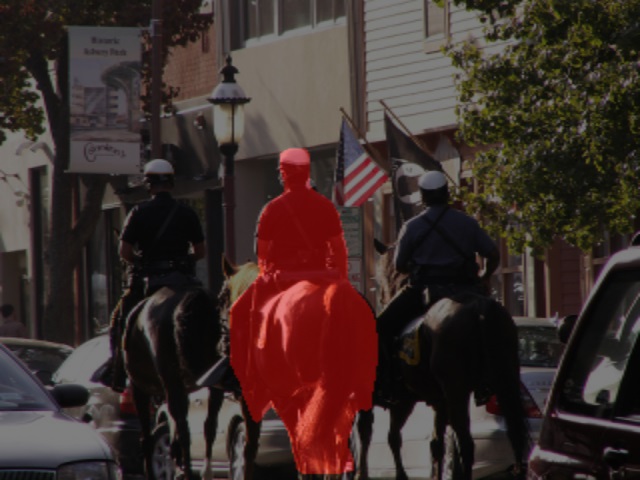}&
\includegraphics[width=0.15\linewidth,height=0.10\linewidth]{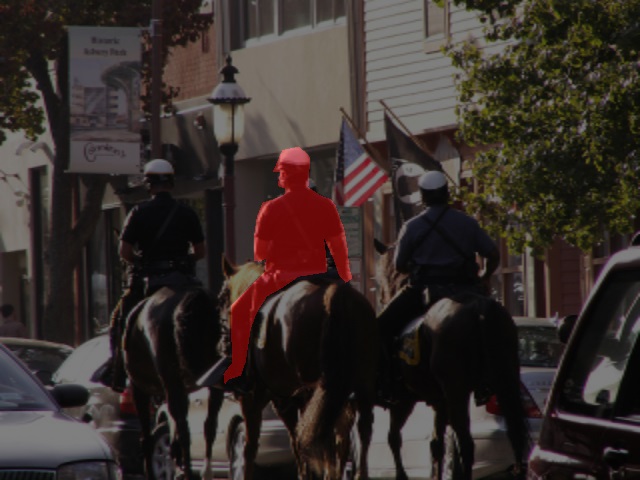}&
\includegraphics[width=0.15\linewidth,height=0.10\linewidth]{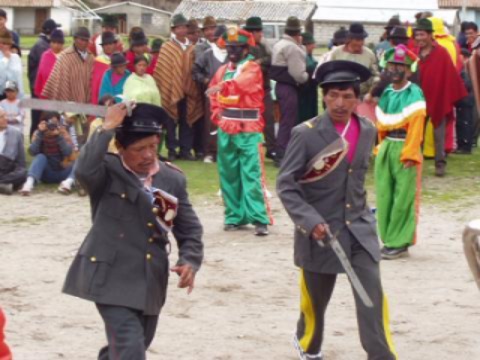}&
\includegraphics[width=0.15\linewidth,height=0.10\linewidth]{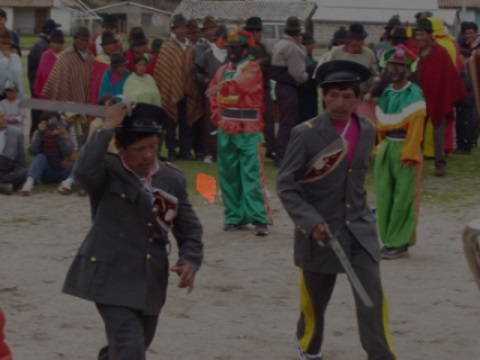}&
\includegraphics[width=0.15\linewidth,height=0.10\linewidth]{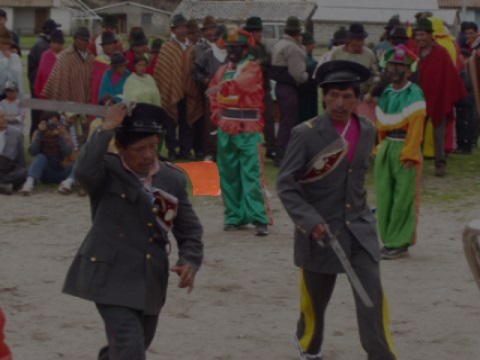}&\vspace{-2.6mm}\\
\fontsize{5pt}{\baselineskip}\selectfont Image&
\fontsize{5pt}{\baselineskip}\selectfont  \makecell[c]{Result}&
\fontsize{5pt}{\baselineskip}\selectfont  \makecell[c]{GT}&
\fontsize{5pt}{\baselineskip}\selectfont Image&
\fontsize{5pt}{\baselineskip}\selectfont  \makecell[c]{Result}&
\fontsize{5pt}{\baselineskip}\selectfont  \makecell[c]{GT}&\vspace{0mm}\\
\end{tabular}
\end{center}
\vspace{-6mm}
\caption{\small Visual examples of the failure cases.}\label{fig:Failure}
\vspace{-3mm}
\end{figure}

We visualize some interesting failure cases in Fig.~\ref{fig:Failure}.
One type of failure occurs when queries are ambiguous. For example, for the top left example, the misspelling of a word ($roght$ $\rightarrow$ $right$) causes part of the semantics of the sentence to be lost. Also, for the top right example, there are two horse butts on the left.
Another case is that when the query contains low-frequency or new words (e.g. in the bottom left example, $cop$ rarely appears in the training data), our method sometimes fails to segment out the core region accurately. This problem may be alleviated by using one/zero-shot learning.
Finally, we observed that sometimes small objects cannot be segmented completely (the bottom right example). This phenomenon can be alleviated by enlarging the scale of input images. Fortunately, the ACM is insensitive to the size (Tab.~\ref{tab:gpumemory} for details).

Through the analysis of successful (Fig.~\ref{fig:performance}) and failure cases, we think that the co-attention module can learn the high-order cross-modal relationships even in some complicated semantic scenarios. It enables the network to pay more attention to the correlated, informative regions, and produce discriminative foreground features.
\section{Conclusion}
In this paper, we propose an encoder fusion network with co-attention embedding (CEFNet) to fuse multi-modal information for referring image segmentation. Compared with the decoder fusion strategy, our strategy adequately utilizes language to guide multi-model feature learning without increasing computational complexity. The designed co-attention module can promote the matching between multi-modal features and strengthen their targeting ability. Moreover, a boundary enhancement module is equipped to make the network pay more attention to the details. Extensive evaluations on four datasets demonstrate that the proposed approach outperforms previous state-of-the-art methods both in performance and speed. In future, we can extend our co-attention module to the one-stage grounding to promote the integration of language and vision.

\section*{Acknowledgements}
This work was supported in part by the National Key R\&D Program of China \#2018AAA0102003, National Natural Science Foundation of China \#61876202, \#61725202, \#61751212 and \#61829102, the Dalian Science and Technology Innovation Foundation \#2019J12GX039, and the Fundamental Research Funds for the Central Universities \#DUT20ZD212.

{\small
\bibliographystyle{ieee_fullname}
\bibliography{egbib}
}

\end{document}